\documentclass{article}

\usepackage[final]{corl_2026} 

\usepackage{amsmath, mathtools, amssymb, amsthm}
\usepackage{siunitx}

\usepackage{hyperref}
\usepackage{caption}
\usepackage{comment}
\usepackage{hang}
\usepackage{wrapfig}
\usepackage{booktabs}

\usepackage{algorithm}
\usepackage[noend]{algpseudocode}

\usepackage[capitalize,nameinlink]{cleveref}
\crefformat{equation}{(#2#1#3)}

\newcommand{\Method}{\texttt{TERRA}}

\newcommand{\R}{\mathbb{R}}

\newcommand{\parens}[1]{\left( #1 \right)}

\newcommand{\norm}[1]{\left\lVert #1 \right\rVert}
\newcommand{\normsmall}[1]{\lVert #1 \rVert}

\newcommand{\dataset}{\mathcal{D}}
\newcommand{\simdataset}{\dataset_\mathtt{sim}}
\newcommand{\realdataset}{\dataset_\mathtt{deploy}}
\newcommand{\simpolicy}{\pi^\mathtt{sim}}
\newcommand{\loss}[1]{\mathcal{L}^\mathtt{#1}}

\newcommand{\task}{\mathtt{task}}
\newcommand{\explore}{\mathtt{explore}}

\title{Explore Where It Matters: Targeted Exploration for Efficient Rover Adaptation}

\author{
\textbf{Jacob Levy}$^{1}$, 
\textbf{Arthur Bouton}$^{2}$, 
Tristan D. Hassler$^{2}$, 
Hari Nayar$^{2}$, 
David Fridovich-Keil$^{1}$ \\
$^1$Department of Aerospace Engineering, University of Texas at Austin\\
$^2$Jet Propulsion Laboratory, California Institute of Technology \\
}

\begin{document}
\maketitle


\begin{abstract}
Future planetary rovers will traverse steep, deformable terrain where slip, sinkage, and terrain deformation can threaten mobility. Although simulation can capture some wheel-soil effects, unmodeled deformable-terrain phenomena create sim-to-real gaps requiring deployment-time adaptation. Yet, rover operations leave little room for extensive trial-and-error. We introduce \Method\ (Targeted Exploration for Reward-Relevant Adaptation), an online adaptation framework that efficiently seeks deployment data exposing task-relevant model errors and corrects them. \Method\ adapts a simulation-trained world model with a reward-weighted prediction loss and trains a loss predictor to estimate where future performance-relevant errors are likely to occur, using this signal to guide exploration during MPC. In simulation, \Method\ reaches approximately 90\% slope-climbing success within five trials under severe out-of-distribution pretraining conditions, outperforming exploration baselines in reward, completion time, and energy to first success. On the ERNEST rover, \Method\ recovers coordinated active-suspension locomotion on a deformable Mars-simulant slope within three to five real-world trials. Project website: \texttt{\url{https://terra-explore.github.io}}

\end{abstract}



\begin{figure}[H]
    \centering
    \includegraphics[width=.99\textwidth]{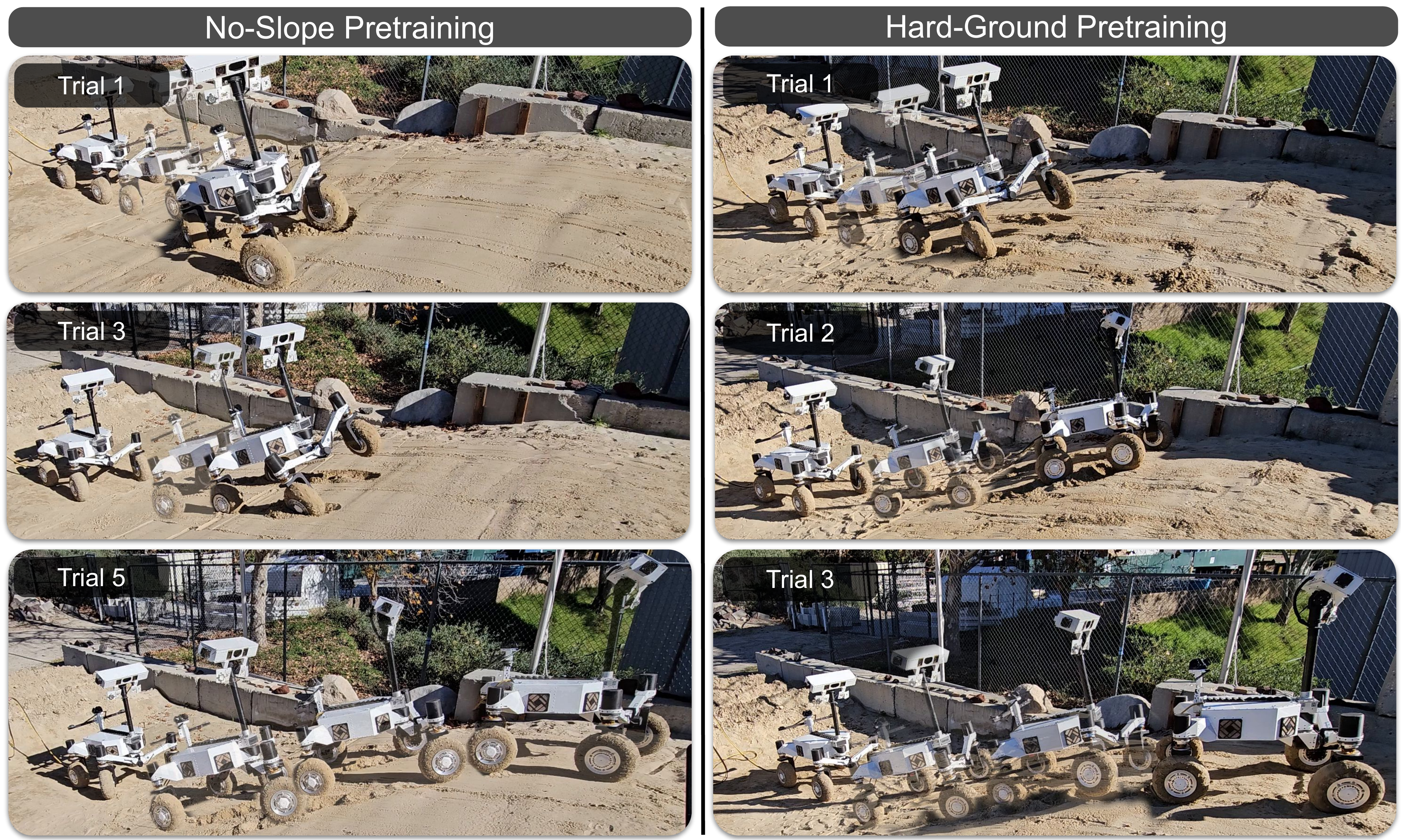}
    \caption{
    \footnotesize{
    \Method\ rapidly adapts to an out-of-distribution deformable slope. No-Slope pretraining never includes slopes, while Hard-Ground never requires coordinated use of the Active Gimbal Suspension (AGS). Both initially entrap the ERNEST rover, but \Method\ learns to climb using its AGS within three to five trials.
    }
    }
    \vspace{-1em}
    \label{fig:runs}
\end{figure}

\section{Introduction}
\label{sec:intro}

\begin{wrapfigure}{r}{0.30\textwidth}
    \centering
    \vspace{-1.3em}
    \includegraphics[width=0.30\textwidth]{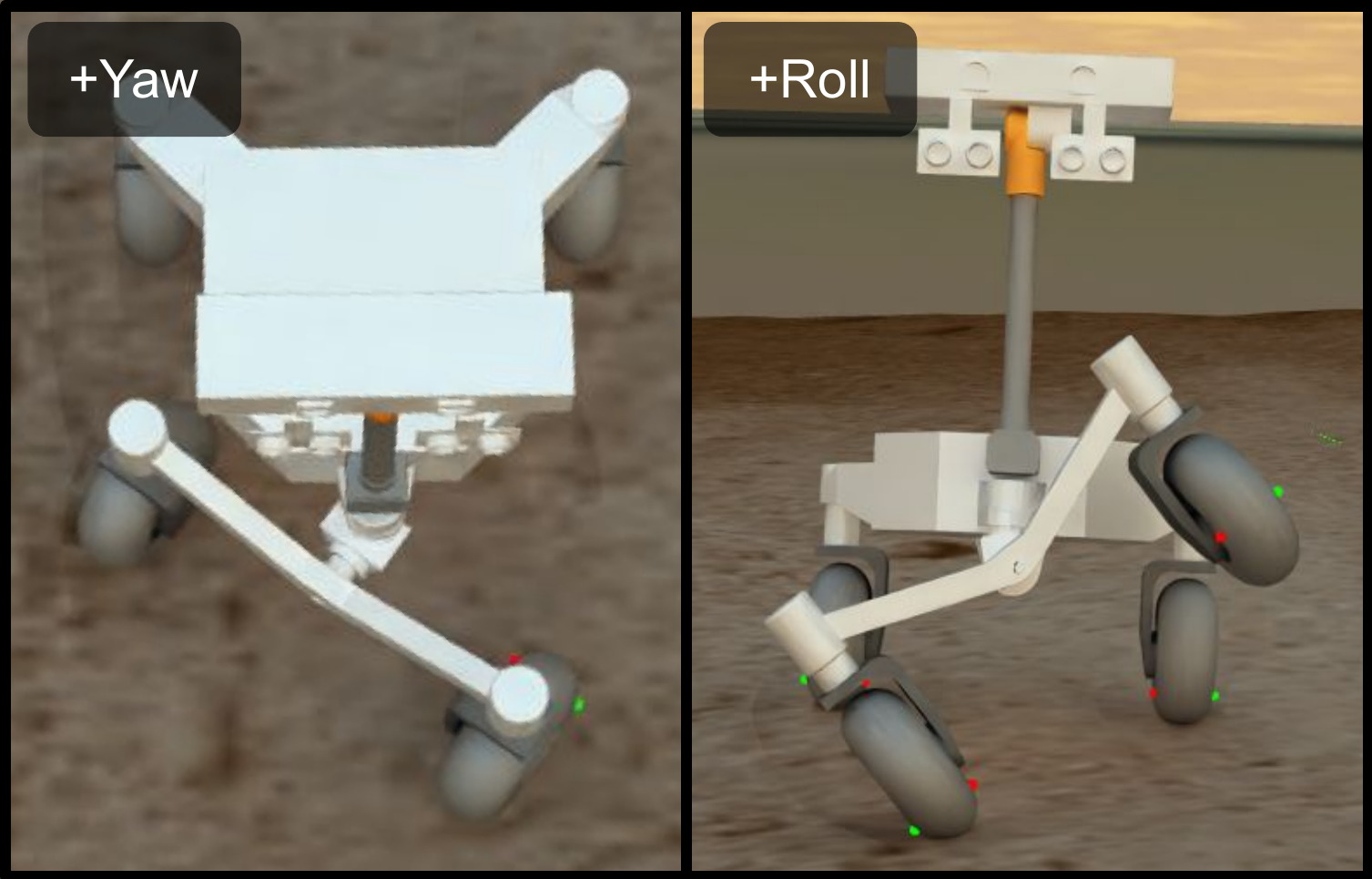}
    \caption{\footnotesize{
    AGS actuation in sim.
    }}
    \label{fig:ags}
\end{wrapfigure}
Future Mars and lunar rover missions are increasingly motivated by science targets in steeper, rougher, and more deformable terrain than previous missions traversed. On Mars, compelling sites tied to habitability, geology, and climate history include mid-latitude regions with steep slope exposures~\cite{helmick2009terrain,quadrelli2015guidance,national2022origins}, while on the Moon, permanently shadowed regions within steep-walled polar craters may host near-surface water ice with scientific and resource value~\cite{tye2015age,li2018direct,chakraborty2024reachability}. These terrains may be covered by loose granular regolith, for which steep slopes can amplify wheel slip and sinkage, increasing the risk of entrapment~\cite{shrivastava2020material}. Such hazards have already affected rover operations: Opportunity suffered substantial slip and sinkage while traversing sandy terrain toward Endeavour Crater~\cite{arvidson2011opportunity}, while Spirit became permanently immobilized in weak soil at Troy, ending the mission~\cite{sanderson2010mars}.

The ERNEST rover (\cref{fig:runs}) is a \SI{75}{\kg}, four-wheeled platform developed to study and address these mobility challenges~\cite{bouton2026learning}. It features a two degree-of-freedom (DoF) Active Gimbal Suspension (AGS) connecting the front bogie to the main chassis. This mechanism has two collocated revolute joints that are actively controlled: a yaw joint that rotates the bogie about a vertical axis relative to the chassis, and a roll joint that rotates it about the bogie’s longitudinal axis (\cref{fig:ags}). Together, these joints enable steering, wheel reconfiguration, active load redistribution, and wheel lifting. Through coordinated actuation, ERNEST can execute crawling-like locomotion: the roll joint alternately shifts load to create anchoring wheels, while the yaw joint repositions the bogie to pull the chassis forward, enabling traversal of soft, steep slopes that can immobilize passive-suspension rovers.

However, effectively using these additional DoFs is challenging because successful locomotion depends on appropriate coordination of the suspension under complex, terrain-dependent wheel-soil interaction. Reinforcement learning (RL) is a natural framework for discovering such strategies in contact-rich systems~\cite{bouton2026learning, levy2025learning,  lee2020learning, kumar2021rma, nahrendra2023dreamwaq}. Yet, because large-scale real-world trial-and-error is infeasible for planetary rovers, policy learning must rely predominantly on simulation. This creates a critical sim-to-real gap: classical terramechanics models capture basic wheel-soil effects such as slip and sinkage, but not soil yielding, flow, excavation, and the resulting terrain deformation during aggressive maneuvers~\cite{bekker1960off, janosi1961analytical, ishigami2007terramechanics}. For example, they cannot capture failure modes in which the rover digs itself into a rut. These effects can substantially alter rover mobility, but high-fidelity simulators that resolve them remain computationally prohibitive for RL-scale training~\cite{tasora2019deformable, xia2011finite, zhang2006simulation, kumar2019scalable}. Consequently, policies trained in simulation inevitably face unmodeled dynamics when deployed in real environments, even when those environments are terrestrial Mars or lunar analogs.

Planetary rovers with articulated mobility mechanisms therefore require not only simulation-trained locomotion strategies, but also mechanisms for rapid deployment-time adaptation to out-of-distribution terrain dynamics at the target site. This adaptation must occur onboard and be highly efficient: rover surface operations are constrained by limited energy, communication delays, bandwidth restrictions, and operational planning workflows, leaving little room for prolonged trial-and-error or human-in-the-loop iteration~\cite{norris2005mars, bresina2005activity, rabideau2020onboard, toupet2025enhanced}. Accordingly, the challenge is not merely to collect more data, but to collect the \emph{right} data: in steep, deformable terrain, not all sources of uncertainty matter equally for control, and the most important ones directly determine whether the rover makes progress or becomes entrapped. In this work, we ask: \emph{how can online adaptation and exploration be coordinated so that limited real-world interaction yields the greatest improvement in rover mobility on previously unseen terrain?}

To answer this question, we build on the paradigm of using a simulation-trained world model for real-time planning and adapting it to the deployment environment, rather than direct policy finetuning~\cite{levy2026simulation}. Our key idea is that adaptation should be guided by control relevance, not prediction error alone. We introduce \Method\ (\textbf{T}argeted \textbf{E}xploration for \textbf{R}eward-\textbf{R}elevant \textbf{A}daptation), a performance-aware online adaptation framework for coordinating exploration and model updates under deployment-time dynamics shift. \Method\ uses a reward-weighted dynamics loss to prioritize model errors with the greatest impact on control performance, and trains a loss predictor to forecast where such errors are likely to occur over the same candidate action sequences evaluated by the planner. This prediction becomes a targeted exploration signal, directing data collection toward interactions where reducing model error should yield the largest task-performance gains. We validate \Method\ through extensive simulated slope-climbing experiments and onboard hardware deployment on ERNEST, demonstrating rapid adaptation to severe out-of-distribution terrain conditions.


\section{Related Work}
\label{sec:related}

\textbf{Model-Based Adaptation.} Model-based RL improves sample efficiency by using learned dynamics for policy improvement though imagined rollouts~\cite{janner2019trust, hafner2019dream, levy2025learning} or by planning through learned models~\cite{hafner2019learning, hansen2022temporal}. In robotics, pretrained models have also been paired with sampling-based MPC and adapted during deployment to compensate for unseen conditions~\cite{levy2026simulation, xiao2025anycar, levy2025meta}. \Method\ builds on this model-based adaptation paradigm, but differs by focusing updates on reward-relevant prediction errors and by actively selecting interactions expected to improve downstream MPC performance.

\textbf{Exploration for Model Learning.} Intrinsic-motivation methods encourage agents to collect informative experience using signals such as information gain~\cite{houthooft2016vime}, prediction error~\cite{schmidhuber1991possibility, pathak2017curiosity}, novelty~\cite{burda2019exploration}, model disagreement~\cite{pathak2019self, sekar_planning_2020}, or reward uncertainty~\cite{liangreward}. \Method\ builds on the self-modeling perspective of~\citet{haber2018learning}, where an agent predicts its own model errors to guide exploration. However, \Method’s loss predictor forecasts reward-weighted dynamics loss over the same candidate action sequences evaluated by MPC, making exploration task-relevant, planner-aligned, and self-regulating as online adaptation corrects those errors.

\textbf{Rover Mobility.} Classical rover-mobility work uses terramechanics-based control to manage slip and sinkage on deformable soil~\cite{helmick2009terrain, ishigami2009slope}. Recent learning-based methods predict slip or traversability from visual terrain cues and onboard experience~\cite{cunningham2017locally, endo2026deep}, while RL with domain randomization can learn active rover locomotion strategies across varied terrain conditions~\cite{bouton2026learning}. \Method\ targets the complementary regime where simulation does not capture the deployment terrain dynamics: rather than relying on traversability prediction or zero-shot policy transfer, it adapts a world model online from very few real trials, with exploration directed toward the model errors that most limit mobility.


\section{Preliminaries and Overview}
\label{sec:prelims}

\textbf{Platform Setup and Simulation Data.} We consider the ERNEST rover~\cite{bouton2026learning}. At each timestep $t$, the robot receives an observation $o_t \in \mathcal{O}$ with proprioception, odometry, wheel forces, and a local terrain height map. The action $a_t \in \mathcal{A} \subset \R^3$ consists of a commanded forward velocity and AGS joint-angle targets, while the corresponding wheel and AGS torques are generated by the low-level controller from~\cite{bouton2026learning}. We define a task reward $r^{\task}_t = R^{\task}(o_t, a_t, o_{t+1})$ that encourages progress along a predefined path while penalizing path deviation and excessive AGS usage. Using the same DARTS simulation environment and reinforcement-learning procedure as~\cite{bouton2026learning}, we train a simulation policy $\simpolicy$ and save tuples $(o_t, a_t)$ encountered during training to a simulation dataset $\simdataset$. The full observation space and reward function is provided in \cref{app:platform}.

\begin{figure}[t]
    \centering
    \includegraphics[width=.99\textwidth]{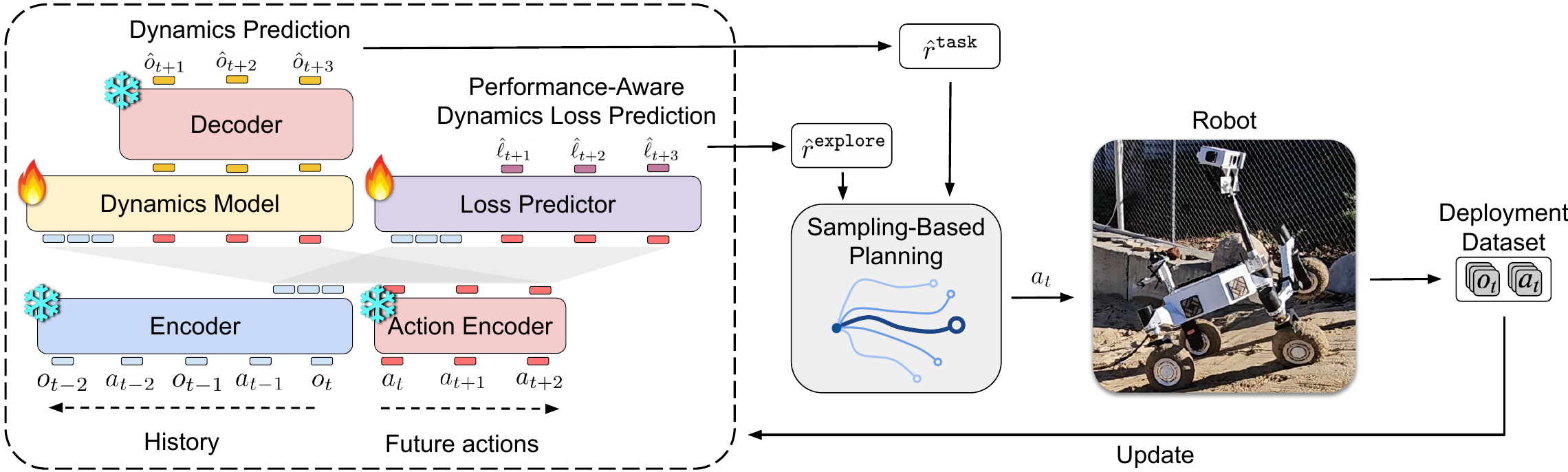}
    \caption{
    \footnotesize{
    \Method\ overview. A simulation-pretrained world model rolls out candidate action sequences for sampling-based planning, while a loss predictor estimates where future performance-relevant dynamics errors are likely to occur. The planner combines predicted task reward with this targeted exploration signal, executes the selected action on the rover, and updates both models with the resulting deployment data.
    }
    }
    \vspace{-1em}
    \label{fig:method}
\end{figure}

\textbf{Model Pretraining and Deployment.} Building on SimDist~\cite{levy2026simulation}, we pretrain a world model in simulation, use it for sampling-based planning in the target environment, and improve performance through online adaptation of the predictive dynamics. Specifically, using $\simdataset$, we pretrain a transformer-based world model of the form:
\begin{equation}
\label{eq:wm}
\begin{array}{l@{\;}ll@{\;}l}
\text{Encoder:} & z_{t-H:t} = E^\mathtt{hist}_\theta(o_{t-H:t}, a_{t-H:t-1}) &
\text{Action Enc.:} & \tilde{a}_{t:t+T-1} = E^\mathtt{act}_\theta(a_{t:t+T-1}) \\
\text{Dynamics:} & \hat{z}_{t+1:t+T} = f_{\theta, \phi}(z_{t-H:t}, \tilde{a}_{t:t+T-1}) &
\text{Decoder:} & \hat{o}_{t+1:t+T} = D_\theta(\hat{z}_{t+1:t+T})  \\
\end{array}
\end{equation}
that conditions on an observation-action history of $H$ timesteps and a future action sequence to predict $T$ future observations. Per SimDist, pretrained parameters $\theta$ remain frozen during deployment and dynamics parameters $\phi$ are finetuned with deployment data. We also train a base policy $\pi_\theta(z_{t-H:t})$ by behavior cloning the simulation policy $\simpolicy$ to provide warm-start action sequences for sampling-based planning. During deployment the MPPI~\cite{williams2016aggressive} planner samples candidate action sequences, rolls them out through the world model \cref{eq:wm}, and scores them using a predicted return; the exact planning objective is specified in \cref{sec:planning}. It then estimates an optimal action sequence by importance-weighting sampled trajectories according to their predicted returns and executes the first action of the selected sequence. As additional deployment data are collected, the parameters $\phi$ of the dynamics model are finetuned online; the specific adaptation procedure is introduced in \cref{sec:adaptation}.


\section{Targeted Exploration for Reward-Relevant Adaptation}
\label{sec:method}

We introduce \Method\ (\cref{fig:method} and \cref{alg:adapt}), a performance-aware framework for coordinating online adaptation and exploration when planning with a simulation-pretrained world model. \Method\ prioritizes prediction errors that most affect the task reward and uses a learned loss predictor to guide planning toward interactions that are informative for improving task performance.

\subsection{Performance-Aware Online Dynamics Adaptation}
\label{sec:adaptation}

The pretrained world model provides a useful prior over rover-terrain interaction, but deployment on out-of-distribution terrain can expose dynamics that are not represented in simulation. Rather than finetuning the entire dynamics model of \cref{eq:wm}, we keep the simulation-trained backbone fixed and adapt only a lightweight set of parameters. We implement this adaptation efficiently using low-rank updates and a reward-weighted prediction loss.

\begin{wrapfigure}{r}{0.51\textwidth}
\vspace{-4em}
\begin{minipage}{0.51\textwidth}
\begin{algorithm}[H]
\caption{
\Method
}
\label{alg:adapt}
\footnotesize
\begin{algorithmic}[1]
\State \textbf{Require:} pretrained parameters $\theta$ from simulation
\State \textbf{Initialize:} $\phi$ and $\psi$
\While{running}
\State \texttt{// Planning}
\State Use MPPI with $f_{\theta,\phi}$, $g_\psi$, and \cref{eq:return} to determine $a_t$
\State Execute $a_t$ in the environment
\State Add $(o_t, a_t)$ to deployment dataset $\realdataset$
\State \texttt{// Adaptation}
\State Sample a batch of segments from $\realdataset$
\State Add the latest segment from $\realdataset$ to the batch
\State Update $\phi$ and $\psi$ with batch using \cref{eq:adaptloss}
\EndWhile
\end{algorithmic}
\end{algorithm}
\end{minipage}
\vspace{-1em}
\end{wrapfigure}
\textbf{Parameter-efficient dynamics adaptation.}
To enable fast onboard updates, we insert low-rank adaptation (LoRA~\cite{hu2022lora}) modules with parameters $\phi$ into the multilayer perceptron (MLP) layers of $f_{\theta,\phi}$, instead of adapting the full transformer. We update the MLPs rather than the attention layers to adapt dynamics predictions while preserving the pretrained structure for attending to latent histories and future action sequences. 
We regularize the LoRA weight updates with $\loss{LoRA}(\phi) = \sum_{i}\normsmall{\Delta W^i_\phi}^2_F$, where $\Delta W^i_\phi$ adapts the $i$-th MLP weight matrix. This regularizer anchors the adapted model near the simulation-pretrained dynamics, mitigating overfitting to limited real-world data and discouraging parameter drift that does not substantially improve prediction on the observed transitions.

\textbf{Performance-aware prediction loss.}
A standard prediction loss encourages the model to reduce observation error uniformly across all dimensions. However, for control, not all errors are equally important: errors in quantities that affect progress, slip, sinkage, or rover pose should be prioritized over weakly relevant dimensions, such as distant height-map cells that are unlikely to affect planning. We therefore adapt the dynamics model with
\begin{equation}
    \label{eq:dynloss}
    \loss{dyn}_t(\phi) = c_1\underbrace{\norm{\hat{o}_{t+1} - o_{t+1}}^2_2}_{\text{ordinary prediction error}} +
    \ c_2 \underbrace{\norm{\nabla_{o_{t+1}}R^{\task}(o_t, a_t, o_{t+1}) \odot \parens{\hat{o}_{t+1} - o_{t+1}}}^2_2}_{\text{reward-weighted prediction error}},
\end{equation}
where $\odot$ denotes elementwise multiplication and $c_1, c_2 > 0$. The second term weights prediction errors by the local sensitivity of the task reward to the next observation, encouraging the model to correct errors most likely to change the planner’s estimated return. We retain the ordinary prediction term to ground the model in the full observation space; otherwise, weakly reward-sensitive dimensions can become underconstrained and drift, degrading reward-relevant predictions.

\subsection{Loss Predictor for Estimating Performance-Aware Loss}

The reward-weighted dynamics loss \cref{eq:dynloss} identifies which observed model errors matter for task performance. 
We also train a \emph{loss predictor} online that the rover uses to predict where such errors are likely to occur in candidate action sequences. As in~\cite{haber2018learning}, we discretize target loss into $N$ bins defined by the thresholds $\{\tau_j\}_{j=1}^N$, treating the problem as one of classification rather than regression.

\textbf{Shared Model Structure.}
Our loss predictor $g_\psi$, with parameters $\psi$, is a transformer that shares the same frozen-encoder inputs as the dynamics model: $\hat{\ell}_{t+1:t+T} = g_\psi(z_{t-H:t}, \tilde{a}_{t:t+T-1})$, where each $\hat{\ell}_k \in \R^N$ contains logits over the discretized loss bins. Because these predictions are conditioned on the same candidate action sequence used for dynamics rollout and task-reward evaluation, they provide an exploration signal aligned with the planner’s rollout-level decision-making.

\textbf{Weighted Classification.}
We train $g_\psi$ with class-weighted cross-entropy ($\mathtt{CE}_w$),
\begin{equation}
    \loss{LP}_t(\psi) = \mathtt{CE}_w\Big( \hat{\ell}_{t+1} , \text{bin}\big(\mathtt{sg}(\mathcal{L}^\mathtt{dyn}_t)\big) \Big),
\end{equation}
where $\text{bin}(\cdot)$ maps the scalar loss into its bin, $\mathtt{sg}$ is the stop-gradient operator which prevents this objective from changing the dynamics model, and $w \in \R^N$ emphasizes high-loss bins. We add this weighting because online losses are highly skewed: familiar interactions produce many small errors, while poorly modeled soil deformation or rut formation can produce large errors. Emphasizing high-loss bins focuses the predictor on rollouts likely to expose reward-relevant model errors, rather than low-error regions with little adaptation value.

\subsection{Exploration-Augmented MPC and Online Adaptation}
\label{sec:planning}

During deployment, \Method\ couples planning and model learning through a single closed loop (\cref{alg:adapt}): MPPI uses the predicted performance-aware loss to choose informative actions, and the resulting data immediately update the dynamics model and loss predictor.

\textbf{Exploration-Augmented Planning.}
For each MPPI candidate, \Method\ rolls out the adapted dynamics model $f_{\theta,\phi}$, evaluates the predicted task reward $\hat{r}^{\task}_k = R^{\task}(\hat{o}_k, a_k, \hat{o}_{k+1})$, and converts the loss-predictor logits into an exploration reward with $\hat{r}^{\explore}_k = \sum_{j=1}^N p_{k+1,j}\tau_j$, where $p_{k+1} = \text{softmax}(\hat{\ell}_{k+1})$. Each candidate action sequence is scored by
\begin{equation}
\label{eq:return}
    \mathcal{R}(a_{t:t+T-1}) = \sum_{k=t}^{t+T-1} \gamma^{k-t}\parens{ \hat{r}^{\task}_k + \lambda \hat{r}^{\explore}_k },
\end{equation}
where $\gamma\in (0,1]$ is the discount factor and $\lambda\geq0$ controls the strength of targeted exploration. Because $\hat{r}^{\explore}_k$ estimates the same reward-weighted dynamics loss used for adaptation, the planner explores interactions expected to expose task-relevant model errors; as those errors are corrected, the bonus decreases and planning naturally shifts toward exploitation.

\textbf{Online adaptation.} After each planning step, the new transition is added to the deployment dataset $\realdataset$. We update the dynamics and loss predictor on trajectory segments of length $H+T+1$ from $\realdataset$, always including the most recent segment so new terrain interactions are incorporated immediately. With $\beta \geq 0$ controlling the LoRA regularization strength, the joint objective is 
\begin{equation}
\label{eq:adaptloss}
    \loss{adapt}_t(\phi, \psi) = \sum_{k=t}^{t+T-1} \parens{ 
 \loss{dyn}_k(\phi) + \loss{LP}_k(\psi)} + \beta\loss{LoRA}(\phi).
\end{equation}



\section{Experiments}
\label{sec:experiments}
We evaluate \Method\ on steep slope traversal under severe pretraining-to-deployment distribution shifts. Our experiments are designed to answer: \textbf{(Q1)} Can \Method\ recover coordinated AGS actuation when the pretrained model fails on the deployment terrain? \textbf{(Q2)} Does targeted exploration via the loss predictor improve adaptation efficiency relative to alternative exploration baselines? \textbf{(Q3)} How does each component of \Method\ contribute to overall mobility performance?

\subsection{Experimental Setup}

\textbf{World Model Pretraining.}
For each condition below, RL follows~\cite{bouton2026learning} using parallel CPU rollouts in the DARTS simulator, and world model pretraining follows~\cite{levy2026simulation} on a single A100 GPU. The world model contains 1.48M frozen pretrained parameters $\theta$ and 328k LoRA dynamics-adaptation parameters, and there are 199k loss-predictor parameters $\psi$. We train with history and prediction horizons $H = T = 10$. Full pretraining and architectural details are provided in \cref{app:details}.

\textbf{Out-of-Distribution Pretraining Conditions.}
We construct two pretraining conditions that each induce a severe train-to-test gap at deployment. In both cases, the soil in the simulator is instantiated with M90 Mars simulant properties~\cite{oravec2021geotechnical}, except as otherwise noted below. Obstacle generation follows~\cite{bouton2026learning}, with domain randomization detailed in \cref{app:details}.

\begin{hangingpar}
1) \textbf{No-Slope.} The rover never sees a slope during pretraining and instead sees smoothed triangular ripples with randomized wavelengths and amplitudes comparable to the rover's wheelbase and wheel height, respectively, and a step consisting of a hard rectangular block with height randomized around wheel height. This exposes the rover to obstacles requiring AGS coordination, but not the sustained slip, sinkage, and load-transfer demands of climbing a deformable slope.
The simulation dataset $\simdataset$ contains 73.7M transitions.
\end{hangingpar}
\begin{hangingpar}
2) \textbf{Hard-Ground.} The rover is pretrained on slopes, but with terrain dynamics that produce hard-ground-like driving behavior. Pretraining environments include: flat terrain with mild Perlin-noise irregularities; the ripples described above; and \ang{15}--\ang{30} slopes, spanning $5$--\SI{10}{\meter} horizontally. To create a high-traction substrate, we set the soil internal friction angle to 60°. This serves as an operational proxy for Hard-Ground traversal: even on the largest ripples and steepest slopes, the rover overcomes obstacles using commanded forward velocity alone, with AGS yaw and roll fixed at zero. Thus, the pretrained model has not learned slope traversal dynamics requiring coordinated AGS actuation under slip and sinkage. $\simdataset$ contains 54.3M transitions.
\end{hangingpar}

\textbf{Evaluation Terrain.}
The evaluation slope is \ang{20} steep, spans \SI{6}{\meter} horizontally with \SI{4}{\meter} lateral bounds. It has parameters matching M90 soil except with internal friction angle reduced to \ang{25}, lowering wheel-soil shear resistance and increasing slip.

\textbf{Evaluation Protocol.}
We evaluate each method over repeated slope-climbing trials. The deployment dataset $\realdataset$ and the updated parameters $\phi$ and $\psi$ are carried forward across trials, so performance measures adaptation over accumulated experience. A trial terminates as a success if the rover reaches the top, and as a failure if it goes out of lateral bounds or does not succeed within $120$ seconds. We run planning and online adaptation at \SI{2}{\hertz}.

\subsection{Simulated Experiments}

\begin{figure}[t]
    \centering
    \includegraphics[width=.99\textwidth]{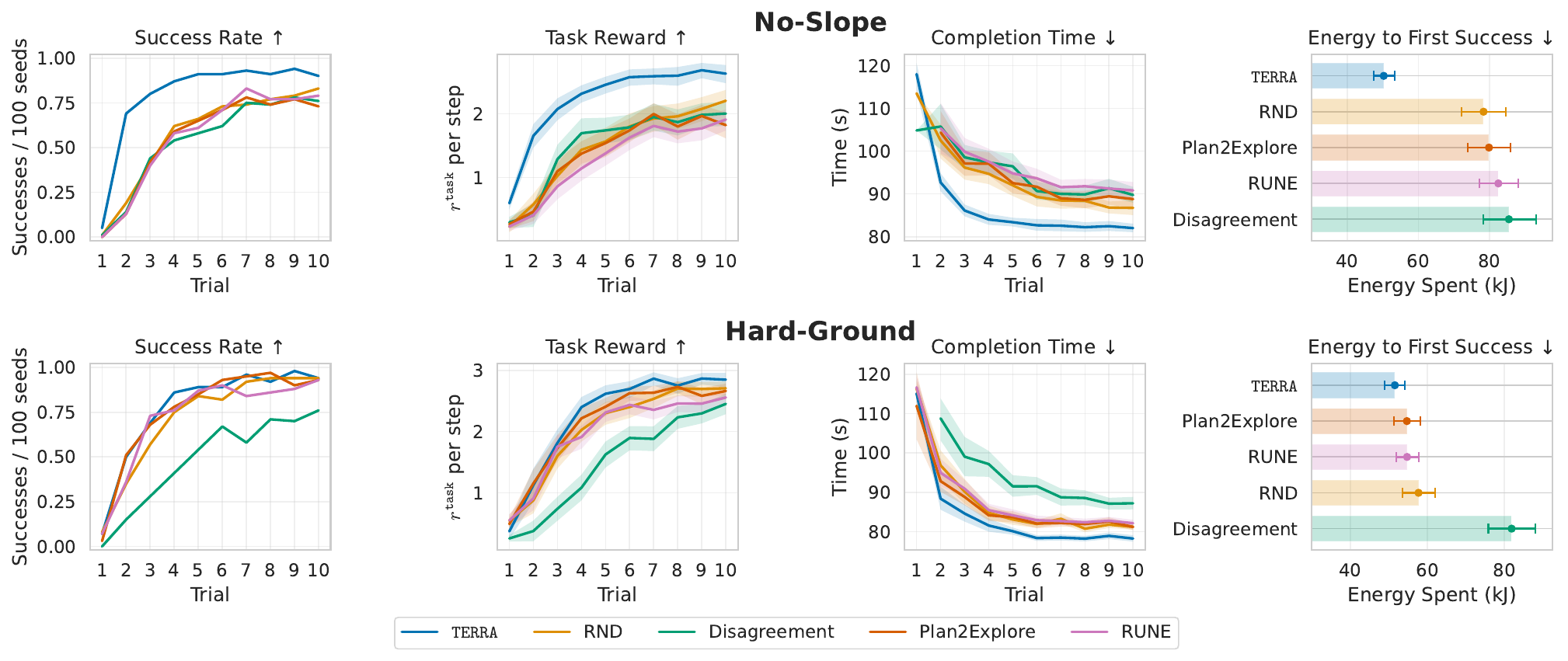}
    \vspace{-0.75em}
    \caption{
    \footnotesize{
    Simulated slope-climbing adaptation under two severe out-of-distribution pretraining conditions. Across 100 random seeds, \Method\ reaches approximately 90\% success within five deployment trials in both No-Slope and Hard-Ground, while achieving higher task reward, shorter completion times, and lower energy to first success than exploration baselines. Curves show means with 95\% bootstrapped confidence intervals. 
    }
    }
    \vspace{-1em}
    \label{fig:simresults}
\end{figure}

\textbf{Results.}
We compare against exploration strategies from: RND \cite{burda2019exploration}, Disagreement \cite{pathak2019self}, Plan2Explore \cite{sekar_planning_2020}, and RUNE \cite{liangreward}; see \cref{app:baselines} for implementation details. Each method is evaluated for 10 deployment trials over 100 random seeds, and \cref{fig:simresults} reports mean metrics with 95\% bootstrapped confidence intervals following \cite{agarwal2021deep}. \Method\ reaches approximately 90\% success within five trials in both No-Slope and Hard-Ground. In No-Slope, \Method\ outperforms all baselines across all reported metrics. In Hard-Ground, several baselines approach \Method’s final success rate, but \Method\ achieves higher reward, shorter completion times, and lower energy to first success (on average). This is especially important for rovers, where energy to first success---the cumulative actuator energy spent before the first successful climb, including failed attempts---captures the operational cost of acquiring competence under limited deployment interaction. Overall, \Method\ is more efficient because it targets mobility-critical model errors rather than generic uncertainty or novelty. We perform additional experiments in \cref{app:exp}.

\begin{figure}[t]
    \centering
    \includegraphics[width=.99\textwidth]{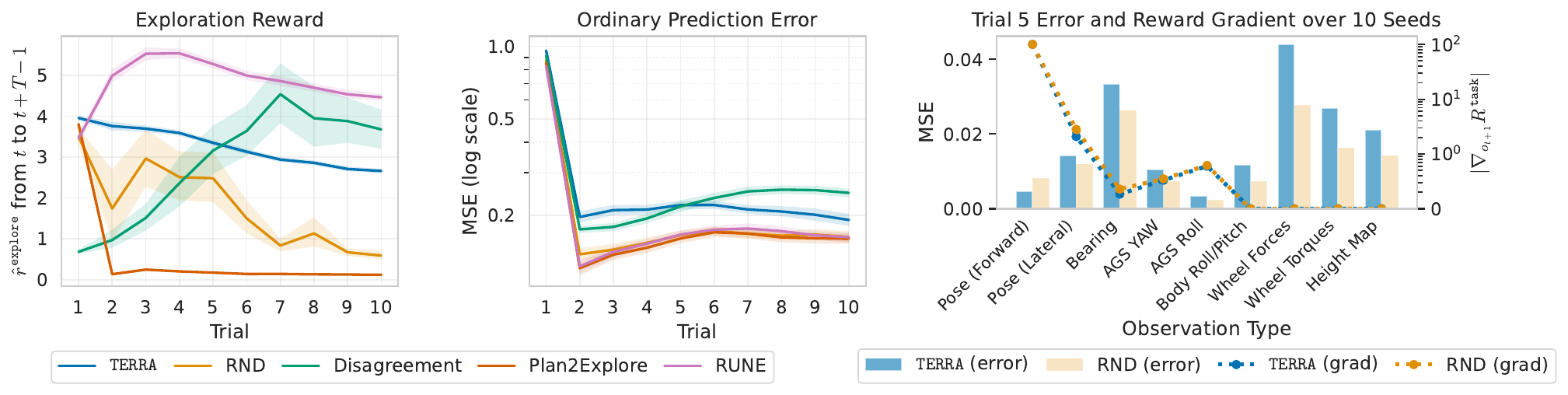}
    \vspace{-0.6em}
    \caption{
    \footnotesize{
    Adaptation metrics for Hard-Ground. Left: each method’s exploration signal evolution over deployment. Middle: lower prediction error alone does not necessarily yield better task performance. Trial-5 error and reward-gradient magnitude for \Method\ and RND, showing that \Method\ reduces error where the reward is most sensitive, while RND achieves lower aggregate error largely on reward-insensitive observation groups.
    }
    }
    \vspace{-1em}
    \label{fig:simanalysis}
\end{figure}

\textbf{Analysis.}
\Cref{fig:simanalysis} explains this efficiency. RND and Plan2Explore exploration rewards drop sharply after the first few trials, suggesting that novelty and representation-disagreement signals are quickly exhausted as the rover visits new states, regardless of whether those states expose errors relevant for climbing. RUNE and Disagreement maintain higher exploration for longer, but spend that exploration less efficiently, yielding smaller reward, completion-time, and energy gains than \Method. In contrast, \Method\ ties exploration directly to the reward-weighted dynamics loss used for adaptation, directing data collection toward model errors whose correction is expected to improve task performance. The middle plot shows that ordinary prediction error is not enough: several baselines achieve lower prediction error than \Method\, yet underperform. The right plot explains why by comparing \Method\ to RND. By Trial 5, averaged over 10 seeds, \Method\ and RND exhibit nearly identical reward-gradient profiles across observation groups, but allocate prediction accuracy differently.RND reduces aggregate error partly in reward-insensitive groups such as wheel forces and height maps, where gradients are zero, while \Method\ achieves lower error on the most reward-sensitive group, forward travel, with only slightly higher error on less reward-sensitive groups.
\begin{wraptable}{r}{0.43\textwidth}
\centering
\vspace{-1.1em}
\caption{
\footnotesize{
Ablations without exploration (expl.) and/or reward-weighted loss (RWL). Across 100 seeds, we report success rate (SR) at trial 5 and energy to first success with 95\% bootstrapped CIs.
}
}
\scriptsize
\setlength{\tabcolsep}{3.5pt}
\begin{tabular}{@{}lcccc@{}}
\toprule
 & \multicolumn{2}{c}{No-Slope} & \multicolumn{2}{c}{Hard-Ground} \\
~ & SR$\uparrow$ & Energy (kJ)$\downarrow$ & SR$\uparrow$  & Energy (kJ)$\downarrow$ \\
\midrule
\Method & \textbf{0.91} & \textbf{50.3 [47, 53]} & \textbf{0.89} & \textbf{51.6 [49, 54]} \\
No expl. & 0.85 & 56.5 [52, 61] & 0.85 & 61.0 [57, 65] \\
No RWL & 0.73 & 80.9 [74, 88] & 0.86 & 60.4 [56, 65] \\
No expl. / RWL & 0.61 & 78.7 [72, 86] & 0.87 & 57.3 [53, 62] \\
\bottomrule
\end{tabular}
\label{tab:ablations}
\vspace{-2em}
\end{wraptable}
Thus, updates should prioritize planning-relevant errors, not overall prediction loss.

\textbf{Ablations.}
\Cref{tab:ablations} shows that \Method’s gains require coupling the reward-weighted dynamics adaptation with exploration. Removing either component degrades performance: exploration must seek reward-relevant errors, and reward weighting helps most when the planner actively collects data that exposes them. The full method performs best because the loss predictor directs interaction toward control-relevant errors, while the reward-weighted loss converts those interactions into task-relevant model updates; see \cref{app:ablations} for additional design ablations.

\subsection{Real-World Validation}

\textbf{Hardware Deployment and Conditions}. 
ERNEST uses AGS joint encoders, stereo-inertial odometry, six-axis force–torque sensors at each wheel assembly, and produces a \SI{10}{\meter}~$\times$~\SI{10}{\meter}, 2.5D elevation map at \SI{1}{\cm} resolution; additional observation details are provided in \cref{app:platform} and~\cite{bouton2026learning}. Planning and online adaptation run onboard on an Intel NUC with an RTX 2060 GPU. The real-world slope mirrors the simulated one, but is \ang{15} steep and constructed from damp M90 Mars simulant. Moisture increases apparent cohesion and preserves wheel tracks under repeated loading, making the hardware setting qualitatively different from the dry simulation and introducing failures such as severe rut formation and entrapment in disturbed soil.

\begin{wrapfigure}{r}{0.40\textwidth}
    \centering
    \vspace{-1.2em}
    \includegraphics[width=.40\textwidth]{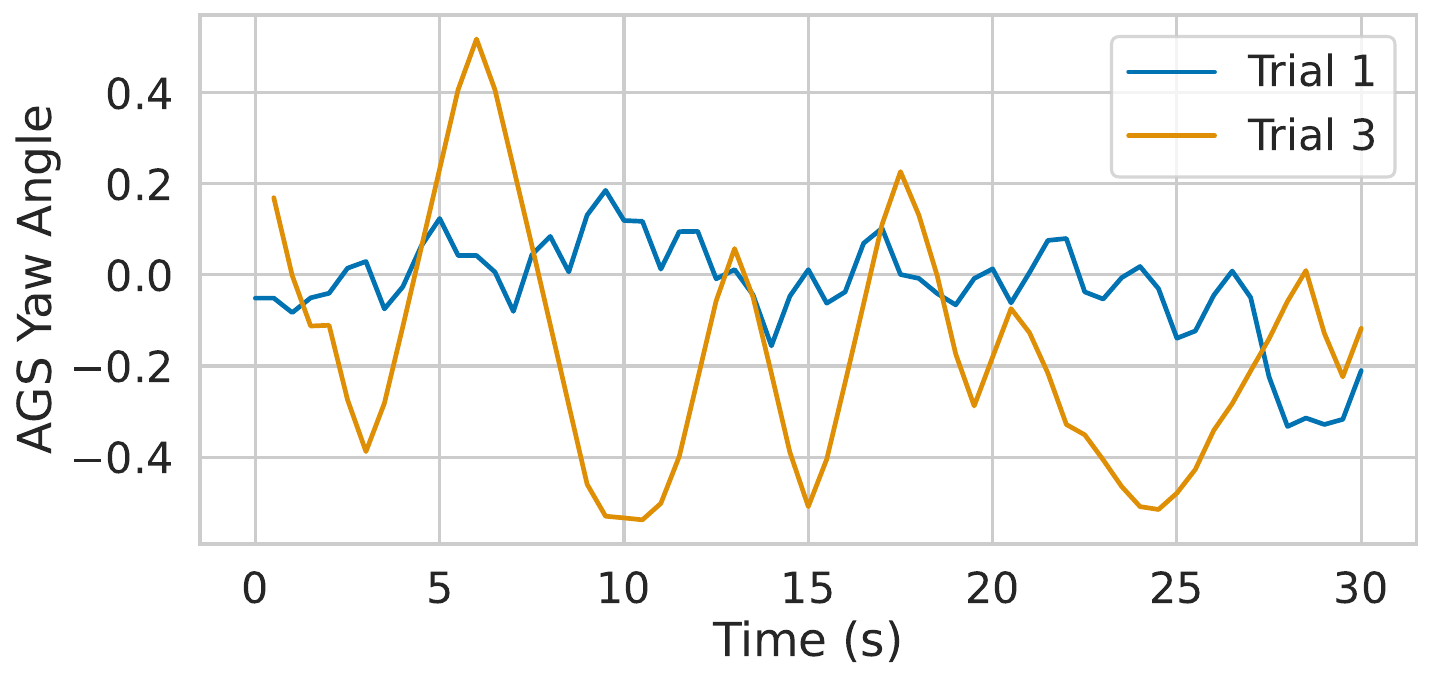}
    \caption{\footnotesize{
    Real-world AGS yaw during Hard-Ground. \Method\ adapts from little AGS use in Trial 1 to large alternating yaw motions by the successful Trial 3 climb.
    }}
    \label{fig:hardyaws}
    \vspace{-1em}
\end{wrapfigure}

\textbf{Results.}
Despite these unmodeled terrain effects, TERRA adapts successfully from both OOD pretraining conditions (\cref{fig:runs}). Before success, failed trials ended either in rut-induced entrapment or out-of-bounds termination; after adaptation, the rover reaches the top by the third trial for Hard-Ground and by the fifth for No-Slope. The Hard-Ground case is especially illustrative because pretraining required little or no AGS actuation. As shown in \cref{fig:hardyaws}, the initial behavior keeps AGS yaw near zero, leading to entrapment, whereas the successful third trial produces large alternating yaw motions consistent with crawling: repositioning and loading the front wheels to pull the chassis uphill. These results show that TERRA can turn a simulation-trained model that fails on an out-of-distribution slope into one that supports successful active-suspension locomotion under real terrain deformation.


\section{Limitations and Future Work}
\label{sec:conclusion}

\Method\ shows that efficient rover adaptation is possible when exploration and model updates are targeted toward the errors that matter most for the task. Across severe simulated distribution shifts and onboard ERNEST deployment, \Method\ converts failed zero-shot slope traversal into successful active-suspension locomotion within a few trials. However, the method assumes that task rewards are available during deployment so that reward-weighted prediction errors can be computed; in robot learning settings, reward functions may rely on privileged training-time information not directly observable onboard. Future work could learn reward models from partial observations or latent states, and use their sensitivities for reward-weighted adaptation and exploration. \Method\ also requires choosing loss-bin thresholds for training the loss predictor, which affects both the supervision target and the scale of the resulting exploration reward. Future work could replace fixed thresholds with adaptive bins that track the online loss distribution.


\clearpage
\acknowledgments{
The authors would like to thank Trey Smith and Brian Coltin for their helpful insights and feedback. The research was carried out at the Jet Propulsion Laboratory, California Institute of Technology, under a contract with the National Aeronautics and Space Administration (80NM0018D0004). This work was supported by a NASA Space Technology Graduate Research Opportunity under award 80NSSC23K1192 and by the National Science Foundation under Grant No. 2409535.
}


\bibliography{refs}  

\newpage
\appendix
\crefalias{section}{appendix}

\section{Platform and Task Reward Function}
\label{app:platform}

\textbf{Platform.} The observation and action spaces are defined in \cref{tab:obs} and \cref{tab:act}. Observations are obtained as in~\cite{bouton2026learning}, except that we use a modified height-map sampling pattern. Using the AGS as the reference frame, we sample longitudinally from \SI{-0.515}{\meter} to \SI{1.185}{\meter} in \SI{0.05}{\meter} increments, and laterally from \SI{-0.6}{\meter} to \SI{0.6}{\meter} in \SI{0.2}{\meter} increments, excluding the centerline at \SI{0.0}{\meter}. This yields three longitudinal columns of height samples in front of each wheel, extending up to \SI{1.185}{\meter} ahead of the AGS.

\begin{table}[h]
\caption{\footnotesize{Observation space.}}
\centering
\scriptsize
\begin{tabular}{lc}
\toprule
Observation & Dimension \\
\midrule
Pose, SE(2) & 3 \\
AGS Yaw and Roll & 2 \\
Body Roll and Pitch & 2\\
Wheel Forces & 12 \\
Wheel Torques & 12 \\
Height Map Sampling & 210 \\
\bottomrule
\end{tabular}
\vspace{0.05cm}
\label{tab:obs}
\end{table}

\begin{table}[h]
\caption{\footnotesize{Action space.}}
\centering
\scriptsize
\begin{tabular}{lc}
\toprule
Action & Dimension \\
\midrule
Commanded Forward Velocity & 1 \\
AGS Yaw Position Command & 1 \\
AGS Roll Position Command & 1\\
\bottomrule
\end{tabular}
\vspace{0.05cm}
\label{tab:act}
\end{table}

\textbf{Task Reward Function.}
The task reward is the weighted sum of each of terms defined in \cref{tab:reward}. 
\begin{table}[h]
\caption{\footnotesize{Task reward function, $R^\mathtt{task}$. Here, $x$ and $y$ are the forward and lateral displacements of the robot; $\beta$ is the deviation from initial robot bearing; and $q_{\mathrm{yaw}}$ and $q_{\mathrm{roll}}$ are the AGS joint angles.}}
\centering
\scriptsize
\begin{tabular}{llc}
\toprule
Term Name & Reward Term & Weight \\
\midrule
Forward advance reward &
\(\displaystyle
\phantom{-}\dfrac{x_{t+1}-x_t}{\Delta t}
\)
& \(50.0\) \\[10pt]

Lateral deviation penalty &
\(\displaystyle -y_{t+1}^{2}\)
& \(2.5\) \\[8pt]

Bearing deviation penalty &
\(\displaystyle -\beta_{t+1}^{2}\)
& \(1.0\) \\[8pt]

AGS Yaw penalty &
\(\displaystyle -q_{\mathrm{yaw},t+1}^{2}\)
& \(1.25\) \\[8pt]

AGS Roll penalty&
\(\displaystyle -q_{\mathrm{roll},t+1}^{2}\)
& \(10.0\) \\[8pt]

Yaw action penalty&
\(\displaystyle -a_{\mathrm{yaw},t}^{2}\)
& \(1.0\) \\[8pt]

Roll action penalty&
\(\displaystyle -a_{\mathrm{roll},t}^{2}\)
& \(1.0\) \\
\bottomrule
\end{tabular}
\vspace{0.05cm}
\label{tab:reward}
\end{table}

\section{Reinforcement Learning, Model, and Pretraining Details}
\label{app:details}

\textbf{Reinforcement Learning.}
RL pretraining for both conditions follows~\cite{bouton2026learning} using TD3~\cite{fujimoto2018addressing} with the hyperparameters listed in \cref{tab:rl_params}. Unlike the original setup, each parallel environment trains an independent policy. During training, we log every observation--action pair $(o_t, a_t)$ encountered across all environments and aggregate these into one simulation dataset $\simdataset$ per pretraining condition. The number of environments used for each condition is given in \cref{tab:num_envs}. Shared domain-randomization parameters are listed in \cref{tab:common_rand}, with No-Slope- and Hard-Ground-specific parameters reported in \cref{tab:no_slope_rand} and \cref{tab:hard_groud_rand}, respectively.

\begin{table}[h]
\caption{\footnotesize{Hyperparameters for RL with TD3.}}
\centering
\scriptsize
    \begin{tabular}{lcc}
        \toprule
        Parameter        & Value  \\
        \midrule
        Discount factor             & 0.99            \\
        Soft target update factor          & \num{5e-3}      \\
        Policy update delay                & 2               \\
        Target policy regularization noise & $\mathcal{N}(0,\, 0.05^2)$ \\
        Bounds of the target policy regularization & 0.2 \\
        \bottomrule
\end{tabular}
\vspace{0.05cm}
\label{tab:rl_params}
\end{table}
\begin{table}[h]
\caption{\footnotesize{Number of parallel environments for each pretraining condition.}}
\centering
\scriptsize
\begin{tabular}{lcccc}
\toprule
Pretraining Condition & Ripples & Step & Flat & Slope \\
\midrule
No-Slope & 1024 & 1024 & -- & -- \\
Hard-Ground & 512 & -- & 512 & 512 \\
\bottomrule
\end{tabular}
\vspace{0.05cm}
\label{tab:num_envs}
\end{table}
\begin{table}[h]
\caption{\footnotesize{Common parameters during RL for both the No-Slope and Hard-Ground pretraining conditions. Individual values indicate fixed parameters and ranges indicate uniform random sampling within the provided ranges.}}
\centering
\scriptsize
    \begin{tabular}{lc}
        \toprule
        Parameter & Value / Range \\
        \midrule

        \multicolumn{2}{l}{\textbf{Initial rover pose}}\\
        Forward position (\unit{\meter}) & $[-1.0,\,0.0]$ \\
        Lateral position (\unit{\meter}) & $[-0.2,\,0.2]$ \\
        Bearing angle (\unit{\degree}) & $[-5.0,\,5.0]$ \\ 
        \addlinespace

        \multicolumn{2}{l}{\textbf{Soil Parameters}}\\
        Soil density (\unit{\kg\per\meter\tothe{3}}) & $1600$ \\
        Cohesion (\unit{\kilo\pascal}) & $[1.8,\,2.2]$\\
        Sinkage exponent & $[1.4,\,2.2]$ \\
        Cohesion modulus ($\mathrm{kN}/\mathrm{m}^{n+1}$) & $[552.1,\,592.1]$\\
        Friction modulus ($\mathrm{kN}/\mathrm{m}^{n+2}$) & $[4815.3,\,5015.3]$\\
        Shear modulus (\unit{\meter}) & $0.0254$ \\
        Coefficients for max normal stress & $0.43,\,0.32$\\
        Coefficients for exit angle & $-0.16,\,0.0$\\
        Slip ratio regularization (\unit{\second\tothe{2}\per\meter\tothe{2}}) & 100 \\ 
        Motion threshold (\unit{\meter\per\second}) & \num{5e-5} \\
        Contact velocity threshold (\unit{\meter\per\second}) & \num{0.01} \\
        \addlinespace

        \multicolumn{2}{l}{\textbf{Ripples}}\\
        Episode length (\unit{\second}) & $60$ \\
        Amplitude (\unit{\meter}) & $[0.15,\,0.35]$ \\
        Wavelength (\unit{\meter}) & $[0.6,\,1.0]$ \\
        Smoothing & $[0.01,\,0.20]$ \\
        Noise amplitude (\unit{\meter}) & $[0.0,\,0.3]$ \\
        
        \bottomrule

    \end{tabular}
\vspace{0.05cm}
\label{tab:common_rand}
\end{table}
\begin{table}[h]
\caption{\footnotesize{Parameters during RL for No-Slope only. Individual values indicate fixed parameters and ranges indicate uniform random sampling within the provided ranges.}}
\centering
\scriptsize
    \begin{tabular}{lc}
        \toprule
        Parameter & Value / Range \\
        \midrule

        \multicolumn{2}{l}{\textbf{Soil Parameters}}\\
        Internal friction angle (\unit{\degree}) & $[30,\,40]$ \\
        \addlinespace

        \multicolumn{2}{l}{\textbf{Step}}\\
        Episode length (\unit{\second}) & $45$ \\
        Height (\unit{\meter}) & $[0.2,\,0.4]$ \\
        Roll (\unit{\degree}) & $[-1,\,1]$ \\
        Pitch (\unit{\degree}) & $[-1,\,1]$ \\ 
        Yaw (\unit{\degree}) & $[-5,\,5]$ \\ 
        Position (\unit{\meter}) & $1.2$ \\ 
        Friction & $[0.5,\,0.9]$ \\

        \bottomrule

    \end{tabular}
\vspace{0.05cm}
\label{tab:no_slope_rand}
\end{table}
\begin{table}[h]
\caption{\footnotesize{Parameters during RL for Hard-Ground only. Individual values indicate fixed parameters and ranges indicate uniform random sampling within the provided ranges.}}
\centering
\scriptsize
    \begin{tabular}{lc}
        \toprule
        Parameter & Value / Range \\
        \midrule

        \multicolumn{2}{l}{\textbf{Soil Parameters}}\\
        Internal friction angle (\unit{\degree}) & $60$ \\
        \addlinespace

        \multicolumn{2}{l}{\textbf{Flat}}\\
        Episode length (\unit{\second}) & $30$ \\
        Roll (\unit{\degree}) & $[-5,\,5]$ \\
        Pitch (\unit{\degree}) & $[-5,\,5]$ \\ 
        Initial rover lateral position (\unit{\meter}) & $[-3,\,3]$ \\
        Initial rover bearing (\unit{\degree}) & $[-90,\,90]$ \\
        Noise amplitude (\unit{\meter}) & $[0.0,\,0.4]$ \\
        \addlinespace

        \multicolumn{2}{l}{\textbf{Slope}}\\
        Episode length (\unit{\second}) & $60$ \\
        Slope angle (\unit{\degree}) & $[15,\,30]$ \\
        $\beta_{\text{start}}$ (\unit{\per\meter}) & $[0.1,\,0.5]$ \\ 
        Length (\unit{\meter}) & $[5,\,10]$ \\ 
        $\beta_{\text{end}}$ (\unit{\per\meter}) & $[0.2,\,2.0]$ \\
        \bottomrule

    \end{tabular}
\vspace{0.05cm}
\label{tab:hard_groud_rand}
\end{table}

\begin{figure}[htb]
    \centering
    \includegraphics[width=.99\textwidth]{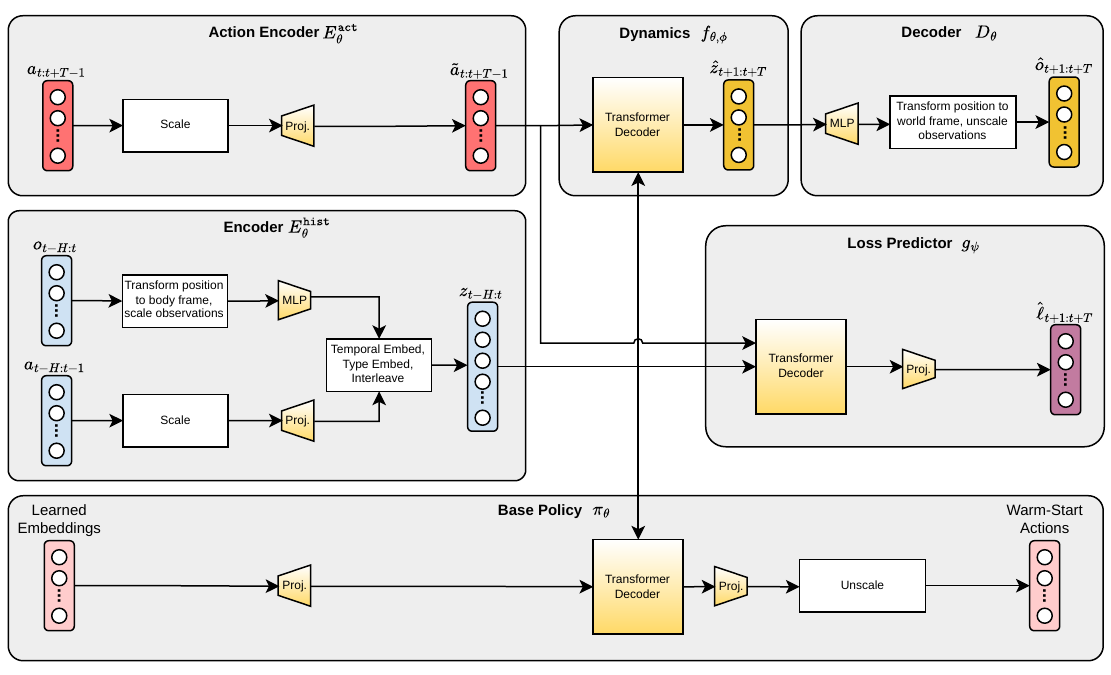}
    \caption{
    \footnotesize{
    Detailed world model and loss predictor architecture.
    }
    }
    \vspace{-1em}
    \label{fig:worldmodel}
\end{figure}

\textbf{World Model and Loss Predictor Structure.}
\Cref{fig:worldmodel} illustrates the world-model, loss-predictor, and base-policy architectures, with corresponding parameters listed in \cref{tab:world_model}. Observation positions are first expressed in the body frame, after which all observation channels are scaled and encoded with an MLP. The action history is separately projected to the embedding dimension. Observation and action embeddings are then temporally embedded, assigned type embeddings to distinguish observations from actions, and interleaved to form the final history encoding. In parallel, the candidate future action sequence is scaled and projected to the same embedding dimension. The history encoding and future-action embeddings are provided to both the dynamics and loss-prediction transformers. The dynamics-transformer outputs are decoded, with predicted positions transformed back to the world frame, to produce future observation predictions. The loss-prediction transformer instead projects its outputs to $\R^N$, yielding logits over the discrete loss bins. Finally, the base-policy transformer uses the history encoding together with learned embeddings to generate warm-start action sequences for MPPI.

\begin{table}[h]
\caption{\footnotesize{World model and loss predictor architectural parameters.}}
\centering
\scriptsize
\begin{tabular}{lc}
\toprule
Parameter & Value \\
\midrule
Embedding dimension & $128$ \\
Encoder (observations) MLP hidden dims & $256, 256$ \\
All transformers MLP hidden size & $512$ \\
Dynamics transformer layers & $4$ \\
Dynamics transformer heads & $8$ \\
Base policy transformer layers & $4$ \\
Base policy transformer heads & $8$ \\
Loss predictor transformer layers & $1$ \\
Loss predictor transformer heads & $4$ \\
Decoder MLP hidden dims & $256, 256$ \\
\bottomrule
\end{tabular}
\vspace{0.05cm}
\label{tab:world_model}
\end{table}

\textbf{Pretraining.}
We pretrain the model parameters $\theta$ on the simulation dataset $\simdataset$ for \num{1.5e6} gradient steps with a batch size of 512. We optimize with Adam using an initial learning rate of \num{2e-4}, cosine decay to \num{2e-5}, and a linear warmup over the first \num{1000} steps. During training, we augment the data by adding zero-mean Gaussian noise to the input observations. Pretraining takes approximately 24 hours on a single NVIDIA A100 GPU.

\section{Implementation Details}

\textbf{MPPI Parameters}
We use MPPI as implemented in TD-MPC \cite{hansen2022temporal} with the hyperparameters listed in \cref{tab:mppi}.
\begin{table}[h]
\caption{\footnotesize{MPPI parameters.}}
\centering
\scriptsize
\begin{tabular}{lc}
\toprule
Parameter & Value \\
\midrule
Candidate actions batch size & $1024$ \\
Noised base policy actions batch size & $2$ \\
Solver iterations & $10$\\
Initial action standard deviation & $[0.2, 0.7, 0.78]$ \\
Minimum action standard deviation & $0.001$ \\
Base policy action standard deviation & $[0.025, 0.05, 0.05]$ \\
Elites & 32 \\
Temperature & 0.25 \\
Momentum & 0.0 \\
Discount & 0.99 \\
\bottomrule
\end{tabular}
\vspace{0.05cm}
\label{tab:mppi}
\end{table}

\textbf{\Method\ Parameters.}
For \Method, we use the parameters listed in \cref{tab:terra_params}.
\begin{table}[h]
\caption{\footnotesize{\Method\ parameters.}}
\centering
\scriptsize
\begin{tabular}{lc}
\toprule
Parameter & Value \\
\midrule
LoRA rank & $64$ \\
Ordinary prediction error weight, $c_1$ & $0.04$ \\
Reward-weighted prediction error weight, $c_2$ & $0.005$ \\
Loss bin thresholds, $\{\tau_j\}_{j=1}^N$ & $\{0, 0.05, 0.1, 0.2, 0.4, 0.8\}$\\
Class weights, $w$ & $[0.5, 0.75, 1, 5, 10, 20]$ \\
Exploration reward strength, $\lambda$ & $0.8$ \\
LoRA regularization strength, $\beta$ & $0.001$ \\
Deployment dataset buffer size & $600$ \\
Batch size & $32$ \\
Learning rate & $0.001$ \\
Steps before updates start & $52$ \\
\bottomrule
\end{tabular}
\vspace{0.05cm}
\label{tab:terra_params}
\end{table}

\section{Baselines}
\label{app:baselines}

We implement the baselines as follows:
\begin{hangingpar}
1) RND \cite{burda2019exploration}: A predictor network $\hat{f}_\psi^\mathtt{RND}$ is trained to predict the output of a frozen network $f^\mathtt{RND}$; exploration reward is $\hat{r}^{\explore}_k = \normsmall{\hat{f}^\mathtt{RND}_\psi(o_{k+1}) - f^\mathtt{RND}(o_{k+1})}^2_2$. For both $\hat{f}_\psi^\mathtt{RND}$ and $f^\mathtt{RND}$, we use an MLP with $2$ hidden layers of size $512$ and an output dimension of $32$.
\end{hangingpar}
\begin{hangingpar}
2) Disagreement \cite{pathak2019self}: An ensemble of $N=5$ LoRA modules is added to the dynamics model, forming the ensemble $\{f^n_{\theta,\phi}\}_{n=1}^N$; exploration reward is \mbox{$\hat{r}^{\explore}_{k+1:k+T} = \text{Var}\{f_{\theta,\phi}^n(h_{k-H:k}, \tilde{a}_{k:k+T-1})\}_{n=1}^N$}.
\end{hangingpar}
\begin{hangingpar}
3) Plan2Explore \cite{sekar_planning_2020}: An ensemble of $N=5$ latent predictors $\{\hat{E}^n_\psi\}_{n=1}^N$ is trained to predict latent future observations $E_\theta(o_{t+1:t+T})$; exploration reward is \mbox{$\hat{r}^{\explore}_{k+1:k+T} = \text{Var}\{\hat{E}_\psi^n(h_{k-H:k}, \tilde{a}_{k:k+T-1})\}_{n=1}^N$}. For $\hat{E}^n_\psi$, we use the same transformer structure as the loss predictor (\cref{app:details}).
\end{hangingpar}
\begin{hangingpar}
4) RUNE \cite{liangreward}: An ensemble of $N=3$ reward predictors $\{\hat{R}^n_\psi\}_{n=1}^N$ is trained to predict $r^{\mathtt{task}}$; exploration reward is $\hat{r}^{\explore}_{k+1:k+T} = \text{Var}\{\hat{R}_\psi^n(h_{k-H:k}, \tilde{a}_{k:k+T-1})\}_{n=1}^N$. For $\hat{R}^n_\psi$, we use the same transformer structure as the loss predictor (\cref{app:details}).
\end{hangingpar}

All baselines use the same exact components as \Method, differing only in exploration signal and, for the main-paper results, adaptation loss. We tune each baseline’s exploration weight $\lambda$ (except Disagreement) to match \Method's Trial-1 exploration reward (\cref{fig:simanalysis}). Because Disagreement begins with near-identical ensemble predictions, we match its mean exploration reward over 10 trials. Thus, exploration incentives are comparable.

Main-paper baselines use standard prediction loss because reward weighting is a \Method\ contribution. For completeness however, we rerun the baselines \emph{with} \Method's reward-weighted adaptation (\cref{tab:full_results}); \Method\ still outperforms across all metrics and has the lowest energy to first success, showing that its targeted exploration is more efficient even when the reward-weighted adaptation mechanism is shared.

\section{Additional Experiments}
\label{app:exp}

\textbf{(E.1) Adding Reward-Weighted Adaptation with Baselines.}
Main-paper baselines use standard prediction loss because reward weighting is a \Method\ contribution. For completeness however, we rerun the baselines \emph{with} \Method's reward-weighted adaptation (\cref{tab:full_results}); \Method\ still outperforms across all metrics and has the lowest energy to first success, showing that its targeted exploration is more efficient even when the reward-weighted adaptation mechanism is shared.

\textbf{(E.2) Robustness to Noise.}
We add per-channel noise sampled from $\mathcal{N}(0, \alpha\sigma^2_i)$, where $\sigma_i$ is each channel’s standard deviation over the simulation dataset. \Method\ remains robust to small/moderate noise and degrades only at large noise (\cref{tab:full_results}), aided by pretraining noise augmentation (\cref{app:details}) and multi-step history/prediction, which favor temporally consistent structure over instantaneous sensor noise.

\textbf{(E.3) Safety.}
Safety can be imposed through the task reward and MPPI constraints. To demonstrate this, we add an experiment to \emph{reduce rover tip-over risk}: a soft penalty begins at $10^\circ$ of body roll angle, while MPPI rejects trajectories predicted to exceed a $15^\circ$ hard limit, \emph{preventing high exploration value from overriding the safety constraint}. Roll-limit violations fall from \textbf{14.2\% to 2.4\%} without sacrificing performance (\cref{tab:full_results}), showing \Method\ can target informative interactions within a safety-constrained feasible set.

\textbf{(E.4) Evaluation on Another Task.}
We add a ripple task under Hard-Ground pretraining. During pretraining, the rover traverses these ripples using forward velocity alone, without AGS actuation. Under the soft-soil deployment dynamics, passive traversal fails and the rover must learn a difficult gait absent from pretraining and \emph{distinct from slope climbing}: lifting a front wheel, placing it on the backside of the next ripple, and pulling forward. \Method\ outperforms the baselines (\cref{tab:full_results}), demonstrating generalization beyond slope climbing.

\section{Additional Ablations}
\label{app:ablations}

\textbf{(F.1) Design Choice Ablations.}
We ablate the following design choices and present the results in \cref{tab:full_results}:
\begin{itemize}
\item \textbf{Removing \Method\ Components.} We remove exploration (expl.) and/or reward-weighted loss (RWL); removing either component degrades performance.
\item \textbf{LoRA rank.} Reducing the LoRA rank from 64 to 32 lowers success and reward in both settings, indicating that \Method\ benefits from the larger low-rank adaptation space.
\item \textbf{LoRA regularization.} Removing LoRA regularization remains competitive, but gives a worse average rank than \Method.
\item \textbf{Full dynamics adaptation.} Adapting the full dynamics model substantially degrades performance, showing that unconstrained online updates are less effective than lightweight LoRA adaptation.
\item \textbf{Ordinary prediction loss.} Removing the ordinary prediction term degrades performance, supporting its role in keeping predictions grounded beyond only reward-sensitive dimensions.
\item \textbf{Class weighting.} Equal class weighting remains competitive, but results in more energy expenditure compared to \Method.
\item \textbf{Latest segment sampling.} Excluding the latest deployment segment from each batch degrades average rank, suggesting that immediately training on the newest interaction helps online adaptation.
\item \textbf{Direct Regression.} We use discretized classification because the goal of the loss predictor is to identify rare high-loss interactions with high adaptation value, rather than precisely estimate their loss magnitudes. Binning reduces sensitivity to extreme loss magnitudes, while class weighting emphasizes these rare high-loss regimes. We ablate this choice by replacing binning with direct regression~\mbox{$\loss{LP}_t(\psi) = (\hat{\mathcal{L}}^\mathtt{dyn}_t - \mathcal{L}^\mathtt{dyn}_t)^2$}, which resulted in worse performance, as shown in \cref{tab:full_results}.
\end{itemize}

\textbf{(F.2) Sensitivity to Hyperparameters}
Scaling \Method's hyperparameters by $0.5\times$ or $2.0\times$ the value from \cref{tab:terra_params} yields similar performance (\cref{tab:full_results}).

\begin{table}[h]
\caption{
\footnotesize
Additional experiments and ablations. Unless noted otherwise, No-Slope, Trial-5, 100 seeds; brackets show 95\% bootstrap CIs.
REL is the geometric mean performance ratio to \Method\ across success rate, reward,
completion time, and energy to first success; $\text{REL}<1$ is worse than \Method\ and $\text{REL}>1$ is better.
}

\centering
\scriptsize
\begin{tabular}{lccccc}
\toprule
~ & SR$\uparrow$ & Reward$\uparrow$ & Time (\unit{\second})$\downarrow$ & Energy (\unit{\kilo\joule})$\downarrow$  & REL$\uparrow$ \\
\midrule

\Method & 0.91 & 2.46 [2.3, 2.6] & 83.4 [82, 84] & 50.3 [47, 53] & 1.00 \\
\addlinespace

\multicolumn{6}{l}{\textbf{(E.1) Adding Reward-Weighted Adaptation to Baselines}}\\
RND & 0.88 & 2.46 [2.3, 2.6] & 84.5 [83, 86] & 53.1 [50, 57] & 0.98 \\
P2E & 0.88 & 2.36 [2.2, 2.5] & 84.8 [83, 86] & 52.4 [50, 55] & 0.97 \\
RUNE & 0.81 & 1.98 [1.8, 2.2] & 87.3 [86, 89] & 52.8 [50, 56] & 0.90 \\
Disagreement & 0.80 & 2.32 [2.1, 2.5] & 90.5 [88, 93] & 59.2 [56, 63] & 0.90 \\

\addlinespace
\multicolumn{6}{l}{\textbf{(E.2) Robustness to Noise}}\\
Add noise,  $\alpha=0.01$ & 0.84 & 2.35 [2.2, 2.5] & 83.7 [83, 85] & 50.8 [48, 54] & 0.97 \\
Add noise, $\alpha=0.05$ & 0.87 & 2.27 [2.1, 2.4] & 86.3 [85, 88] & 61.0 [57, 65] & 0.92 \\
Add noise, $\alpha=0.10$ & 0.57 & 1.45 [1.2, 1.7] & 93.7 [92, 96] & 67.8 [63, 73] & 0.70 \\

\addlinespace
\multicolumn{6}{l}{\textbf{(E.3) Safety}}\\
$15^\circ$ body roll limit & 0.88 & 2.47 [2.3, 2.6] & 83.8 [83, 85] & 48.6 [46, 51] & 1.00 \\

\addlinespace
\multicolumn{6}{l}{\textbf{(E.4) Evaluation on Ripples (Hard-Ground, 25 random seeds, Trial-10)}}\\
\Method & 0.28 & 3.78 [3.4, 4.1] & 55.9 [54, 59] & 46.9 [38, 57] & 1.00 \\
RND & 0.16 & 3.98 [3.7, 4.2] & 54.0 [50, 61] & 57.1 [47, 67] & 0.85 \\
Disagreement & 0.00 & 2.91 [2.5, 3.3] & No successes & No successes & N/A \\
Plan2Explore & 0.16 & 4.13 [3.9, 4.4] & 56.2 [51, 61] & 54.0 [44, 64] & 0.86 \\
RUNE & 0.16 & 3.59 [3.3, 3.9] & 59.9 [58, 63] & 49.2 [36, 63] & 0.83 \\

\addlinespace
\multicolumn{6}{l}{\textbf{(F.1) Design Choice Ablations.}}\\
No expl. & 0.86 & 2.39 [2.2, 2.5] & 85.5 [84, 87] & 56.5 [52, 61] & 0.94 \\
No RWL & 0.73 & 1.76 [1.5, 2.0] & 90.2 [88, 92] & 80.9 [74, 88] & 0.76 \\
No expl / RWL & 0.61 & 1.49 [1.3, 1.7] & 90.1 [88, 93] & 78.7 [72, 86] & 0.70 \\
LoRA rank of 32 & 0.83 & 2.38 [2.2, 2.5] & 83.2 [82, 85] & 51.9 [49, 55] & 0.96 \\
No LoRA regularization & 0.85 & 2.39 [2.2, 2.5] & 83.3 [82, 84] & 51.0 [48, 54] & 0.97 \\
Adapt full dynamics model & 0.68 & 1.54 [1.3, 1.7] & 97.4 [95, 100] & 99.9 [94, 106] & 0.67 \\
No ordinary prediction loss & 0.73 & 2.12 [2.0, 2.3] & 86.5 [85, 88] & 60.2 [57, 64] & 0.86 \\
Equal class weighting & 0.94 & 2.55 [2.4, 2.7] & 83.4 [82, 85] & 53.4 [50, 58] & 1.00 \\
Latest segment not in batch & 0.82 & 2.29 [2.1, 2.4] & 84.0 [83, 85] & 53.8 [50, 58] & 0.94 \\
Direct regression & 0.82 & 1.67 [1.5, 1.9] & 93.2 [91, 96] & 61.7 [58, 65] & 0.82 \\

\addlinespace
\multicolumn{6}{l}{\textbf{(F.2) Sensitivity to Hyperparameters}}\\
$0.5\times$ exploration, $\lambda$ & 0.88 & 2.54 [2.4, 2.7] & 82.9 [82, 84] & 51.6 [49, 55] & 0.99 \\
$2.0\times$ exploration, $\lambda$ & 0.91 & 2.40 [2.2, 2.5] & 83.7 [82, 85] & 52.9 [49, 57] & 0.98 \\
$0.5\times$ loss bin thresholds & 0.89 & 2.61 [2.5, 2.7] & 82.4 [81, 84] & 50.8 [48, 54] & 1.01 \\
$2.0\times$ loss bin thresholds & 0.86 & 2.36 [2.2, 2.5] & 83.0 [82, 84] & 56.0 [53, 59] & 0.95 \\
$0.5\times$ reward weighting & 0.92 & 2.46 [2.3, 2.6] & 83.4 [82, 85] & 53.5 [51, 57] & 0.99 \\
$2.0\times$ reward weighting & 0.88 & 2.50 [2.3, 2.6] & 82.8 [82, 84] & 52.3 [49, 56] & 0.99 \\

\bottomrule
\end{tabular}
\label{tab:full_results}
\end{table}

\end{document}